\documentclass[letterpaper, 10 pt, conference]{ieeeconf}  
\IEEEoverridecommandlockouts
\usepackage{graphics} 
\usepackage{epsfig} 
\usepackage{mathptmx} 
\usepackage{times} 
\usepackage{amsmath} 
\usepackage{amssymb}  

\usepackage{multirow}
\usepackage[table,xcdraw]{xcolor}
\usepackage{threeparttable}
\usepackage{eucal}
\usepackage{balance}
\usepackage{mathrsfs}  
\usepackage{comment}
\usepackage[noend]{algpseudocode}
\usepackage{float}
\usepackage{textcomp}
\usepackage{mathcomp}
\usepackage{bm}
\usepackage{cite} 
\usepackage{booktabs}
\usepackage{url}
\usepackage{hyperref}
\usepackage{comment}
\usepackage[whole]{bxcjkjatype} 

\usepackage[hang,small,bf]{caption}
\usepackage[subrefformat=parens]{subcaption}
\usepackage{algorithm}
\usepackage{algpseudocode}
\usepackage{float} 
\usepackage{here}  

\newcommand{\aoki}[1]{}
\newcommand{\honda}[1]{}
\newcommand{\okuda}[1]{}
\newcommand{\suzuki}[1]{}
\newcommand{\important}[1]{}
\newcommand{\considerhere}[1]{}
\newcommand{\trivial}[1]{}

\def\eg{\textit{e.g}\onedot} 
\def\ie{\textit{i.e}\onedot}

\def\eg{{\it e.g.}}

\def\ie{{\it i.e.}}

\title{\LARGE \bf
Switching Sampling Space of Model Predictive Path-Integral Controller \\ to Balance Efficiency and Safety in 4WIDS Vehicle Navigation 
}

\author{Mizuho Aoki$^{1}$, Kohei Honda$^{1}$, Hiroyuki Okuda$^{1}$, and Tatsuya Suzuki$^{1}$
\thanks{*This work was not supported by any organization.}
\thanks{$^{1}$The Department of Mechanical Systems Engineering, Graduate School of Engineering, Nagoya University, Furo-cho, Chikusa-ku, Nagoya, Aichi, Japan, {\tt\small mizuhoaoki1998@gmail.com}}%
}

\begin{document}
\onecolumn

2024 IEEE/RSJ International Conference on Intelligent Robots and Systems (IROS 2024) \\

Accepted paper. Accepted June 2024. \\

\copyright \ 2024 IEEE. Personal use of this material is permitted. Permission from IEEE must be obtained for all other uses, in any current or future media, including reprinting/republishing this material for advertising or promotional purposes, creating new collective works, for resale or redistribution to servers or lists, or reuse of any copyrighted component of this work in other works.

\cleardoublepage
\twocolumn

\maketitle
\thispagestyle{empty}
\pagestyle{empty}

\begin{abstract}

Four-wheel independent drive and steering vehicle (4WIDS Vehicle, Swerve Drive Robot) 
has the ability to move in any direction by its eight degrees of freedom (DoF) control inputs.
Although the high maneuverability enables efficient navigation in narrow spaces,
obtaining the optimal command is challenging due to the high dimension of the solution space.
This paper presents a navigation architecture using the Model Predictive Path Integral (MPPI) control algorithm
to avoid collisions with obstacles of any shape and reach a goal point. 
The key idea to make the problem easier is
to explore the optimal control input in a reasonably reduced dimension that is adequate for navigation.
Through evaluation in simulation, we found that the selecting sampling space of MPPI greatly affects navigation performance.
In addition, our proposed controller which switches multiple sampling spaces according to the real-time situation
can achieve balanced behavior between efficiency and safety.
Source code is available at \url{https://github.com/MizuhoAOKI/mppi_swerve_drive_ros}.

\end{abstract}

\section{INTRODUCTION}

Four-wheel independent drive and steering (4WIDS) vehicles have the potential to enhance the level of vehicle motion as a next-generation power transmission system for automobiles \cite{4wids_survey}\cite{wheel_robot_survey}.
4WIDS is capable of independently controlling the steering angle and drive torque of all four wheels, enabling holonomic movements such as pivoting in place and diagonal movement. 
In particular, unlike other types of holonomic robots (\eg, omnidirectional wheels and mecanum wheels robots), 
steerable vehicles can drive on rough terrain and maintain high-speed stability \cite{wheel_robot_survey, mpc_4wids_1, mppi_3wids_add_2}.
However, due to its high-dimensional input space (\ie, eight degrees of freedom), control can be challenging. In addition, smooth steering control is required to avoid mechanical failures.

In general, gradient-based Model Predictive Control (MPC) is one of the effective approaches for redundant systems, where the input dimension is larger than the state dimension \cite{mpc_4wids_1, mpc_4wids_2, mpc_4wids_3}.
However, gradient-based MPC has challenges such as the inability to handle non-differentiable cost functions (\eg, cost map) or the possibility of converging to local minima in nonlinear problems.
Therefore, sample-based MPC approaches that do not rely on gradients, such as the Model Predictive Path Integral control (MPPI)~\cite{williams2018information}, 
are practically promising and used for a wide variety of applications 
\cite{mppi_1, mppi_2, mppi_3, mppi_4, mppi_5, mppi_6, mppi_7, arruda2017uncertainty, liu2023collision, howell2022predictive}.

\begin{figure}[t]
    \centering
    \includegraphics[width=0.95\linewidth]{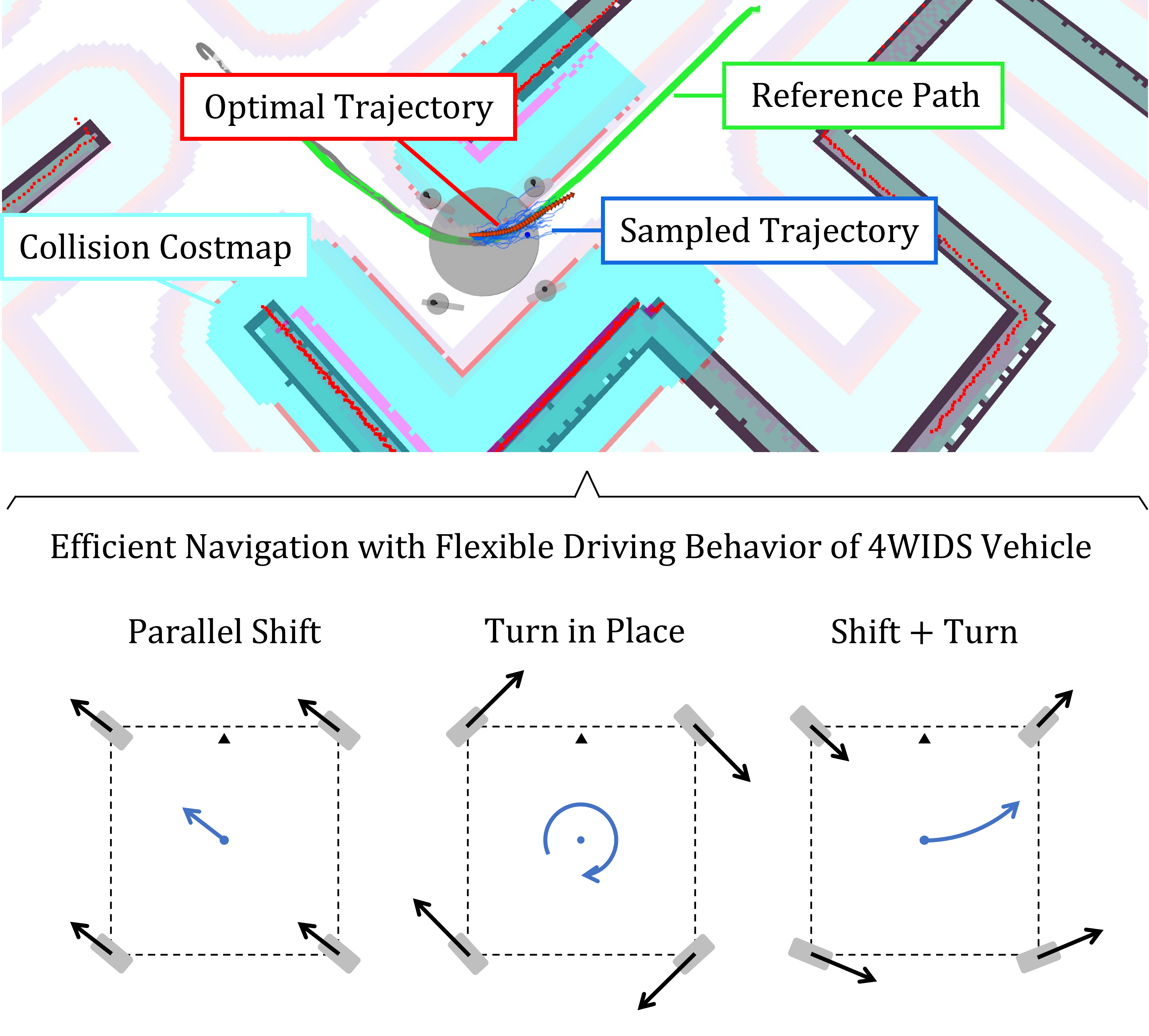}
   \vspace{+0mm}
      \caption{
        4WIDS vehicles can achieve various types of motion, such as moving diagonally and turning in place.
        Our navigation architecture makes good use of these capabilities to achieve efficient and stable navigation in narrow spaces avoiding obstacles.
      }
   \vspace{-7mm}
    \label{fig:top_page_fig}
  \end{figure}

When applying MPPI to redundant systems, the dimension of the input space is a critical issue. The larger the dimension of the input space, the more pronounced the curse of dimensionality becomes. 
This means that the number of samples required to adequately cover the input space increases exponentially, resulting in a significant decrease in sampling efficiency.
How to avoid the curse of dimensionality and apply MPPI efficiently remains a critical issue in controlling redundant systems.

In this paper, we focus on analytically constraining the input space of 4WIDS through geometric constraints, allowing for the application of MPPI to a reduced-dimensional input space. 
As a result, we can avoid the aforementioned curse of dimensionality and maintain the sample efficiency of MPPI while controlling 4WIDS. To the best of our knowledge, this is the first study to apply MPPI to 4WIDS.

The primary contribution of this work is three-fold.
First, we derive that it is possible to reduce the original eight-dimensional input space of 4WIDS to three dimensions by applying static geometrical constraints. 
However, our experiments revealed an insight that the three-dimensional control input space degrades the stability of navigation. 

Therefore, as a second contribution, we demonstrate that \emph{controlling 4WIDS is more effective with a slightly redundant input space than with the bare minimum of three dimensions}. 
Specifically, we opted to apply MPPI to a four-dimensional input space that includes redundancy, instead of sticking to the minimal three dimensions. 
As a result, our empirical simulation results proved to enhance the success rate of navigation tasks.

Third, as a final contribution, we propose an approach to switch the control input space in real-time, depending on the situation.
The hybrid sampling showed balanced performance to achieve both efficiency and a high success rate in navigation tasks.

\section{RELATED WORK}
\subsection{Conventional Control Approaches for 4WIDS}
A simple way to control a 4WIDS vehicle is by limiting its degrees of freedom with adding constraints. 
In \cite{purepursuit_4wids_1}, a constraint is added that the front steering angle is the negative of the rear steering angle and applied pure-pursuit algorithm for path tracking.
This approach is easy to understand and computationally efficient, but it is not suitable for utilizing the vehicle's maneuverability.

Fuzzy logic is a powerful tool to handle full vehicle input space by preparing multiple driving modes and switching them according to the situation\cite{fuzzy_4wids}.
However, it requires a high level of domain knowledge to design each mode and the predefined rules are difficult to deal with unknown environments.

To guarantee the stability of the system in any situation, the Lyapunov function is used to design the control law\cite{lyapunov_4wids, kine_dyna_4wids}.
Stable path tracking can be achieved, 
but considering other tasks such as smooth steering and obstacle avoidance is challenging for the framework. 

Since considering various goals and constraints can broaden the scope of applications, 
Model Predictive Control (MPC) has been widely used, 
which predicts the future state of the target system and gets the optimal control input minimizing a cost function \cite{mpc_4wids_1, mpc_4wids_2, mpc_4wids_3}.
Despite its high flexibility, MPC is computationally expensive. 
To achieve real-time control, using a gradiant-based optimization solver,  linear approximation of the vehicle model,
or assuming an accurate reference trajectory are practical strategies.

However, to perform practical local planning that smoothly avoids obstacles of arbitrary shape,
complex optimization problems including non-convex and non-differentiable formulation are essential, which is difficult to handle with conventional methods.
This paper introduces a sampling-based solver to expand the scope of MPC-based approaches
and aims to achieve more general navigation tasks in cluttered environments.

\subsection{Applications of Sampling-based MPC for Redundant Systems}

Sampling-based MPC is a powerful tool to control redundant systems and has been applied to various complex control problems.

The simplest algorithm is to sample several control sequences and evaluate them to find the best one.
Complex systems such as humanoid robots and quadrupedal robots can be controlled by a simple algorithm \cite{howell2022predictive}.
However, this approach has low sample efficiency and the behavior is likely to be jerky.

Using the MPPI algorithm can enhance the sample efficiency and generate smoother behavior.
Manipulators with high degrees of freedom can successfully avoid obstacles\cite{liu2023collision}, 
and stable manipulation is achieved even when the system model is uncertain\cite{arruda2017uncertainty}.
Combination with reinforcement learning has also been studied\cite{Hansen2022tdmpc}.
Learning environment dynamics and exploring the solution in the latent space successfully operated a 38-DoF control target.

While the application of MPPI is expanding, there is still little knowledge about how to choose the sampling space to explore the solution effectively.
This paper aims to provide a new insight into the sampling space selection with comparative experiments and theoretical analysis.

\section{TARGET TASK DESCRIPTION} \label{sec:target_task}
\begin{figure}[H]
    \centering
    \includegraphics[width=0.8\linewidth]{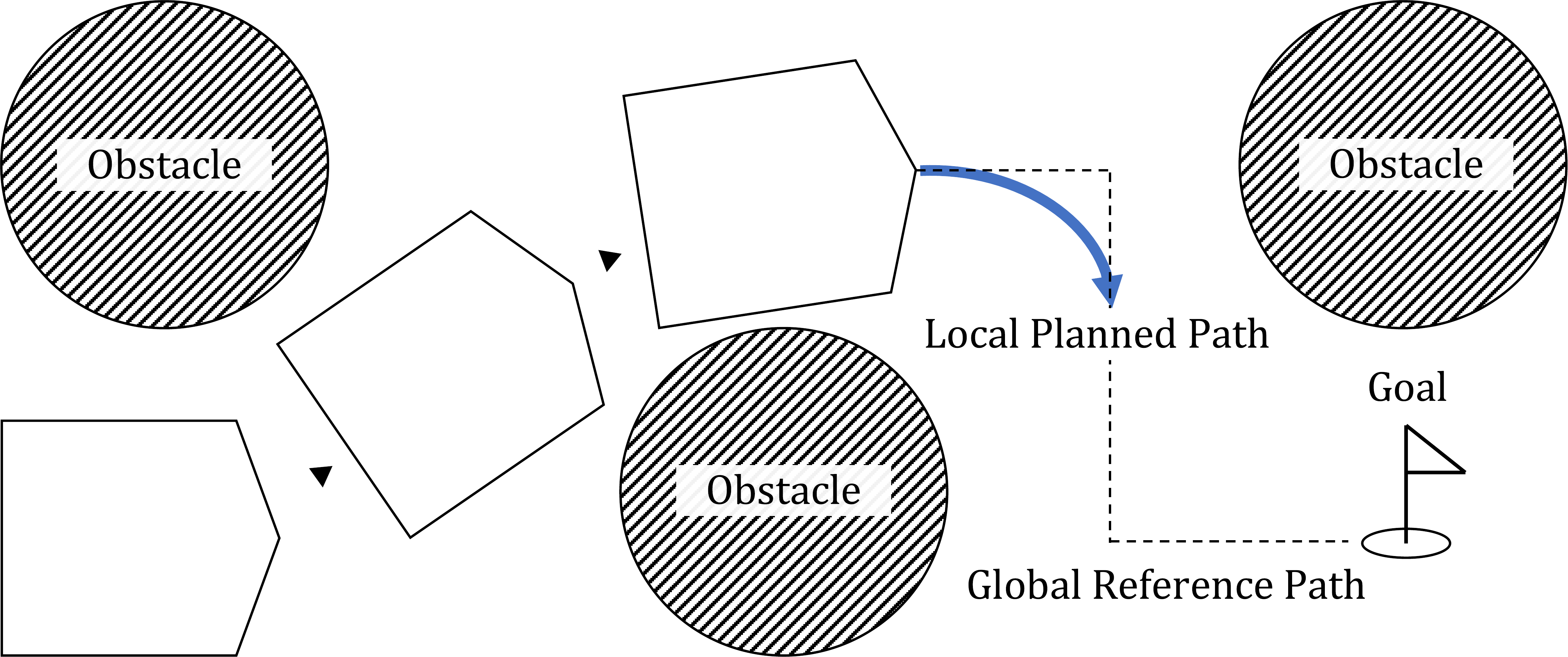}
    \caption{Navigation task}
    \label{fig:target_task}
\end{figure}

The goal of our system is to navigate a 4WIDS vehicle to a given goal while avoiding collision with surrounding obstacles (Fig. \ref{fig:target_task}).
We assume that a global planner generates a reference trajectory, which is a sequence of positions and orientations in a 2D plane.
Therefore, the main focus of our study is to design a local planner that generates a vehicle motion to follow the reference trajectory while safely avoiding obstacles.
The local planner sends an eight DoF vehicle command consisting of a wheel speed and a steering angle for each wheel to the vehicle actuators.
Especially in our navigation environment where the obstacles are densely distributed, the high maneuverability of 4WIDS vehicles, such as small turning radius and diagonal movement, is key to achieving smooth and efficient driving.

\section{Modeling of 4WIDS Vehicle Motion} \label{sec:dynamics}

To achieve accurate vehicle control, it is important to understand the characteristics of vehicle motion. 
In this section, we formulate the model of 4WIDS vehicle behavior, mainly used for sampling future vehicle poses in our model predictive controller.

\begin{figure*}[t]
  \centering
  \includegraphics[width=0.99\linewidth]{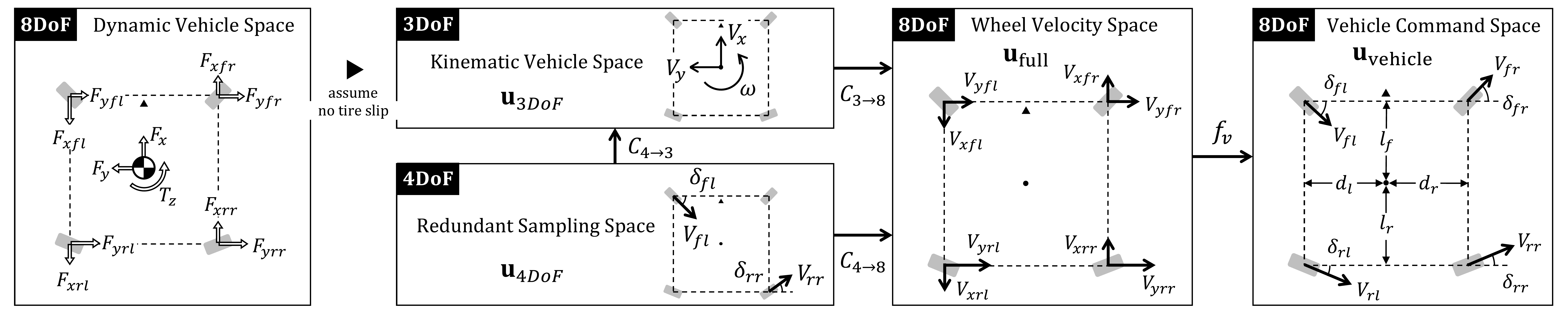}
 \vspace{+0mm}
    \caption{Relationship between spaces; Dynamics(8DoF) can be simplified to Kinematics(3DoF) with assuming no tire slip.
    Two types of sampling space are tested in this study; $\mathbf{u}_{\rm{3DoF}}$ and $\mathbf{u}_{\rm{4DoF}}$ to compare the navigation performance.
    Both spaces can be converted to the vehicle command space $\mathbf{u}_{\rm{vehicle}}$ with conversion matrix $C_{n\rightarrow8} ( n \in \{3, 4\} )$ and nonlinear projection $f_v$.
    }
 \vspace{+0mm}
  \label{fig:space_relationship}
\end{figure*}

\subsection{4WIDS Vehicle Dynamics} 

Although it is a straightforward approach to model vehicle behavior using Newton's laws of motion,
our study does not focus on this approach.
To formulate the full vehicle dynamics, it is necessary to consider the eight tire forces generated by the four wheels, as shown in Fig.\ref{fig:space_relationship}.
However, to obtain the forces, it is necessary to observe tire slip angles, tire slip ratios, and tire physical parameters,
which are not easy to measure accurately in real-world applications \cite{vehicle_model_rajamani}.
Even if we have true dynamics and explore the full eight-dimensional input space, 
most of the solutions will include large tire slips, which cause unstable vehicle behavior.
Therefore, predicting vehicle motion with full dynamics is not always practical, especially for real-world applications.  

\subsection{4WIDS Vehicle Kinematics} 
An efficient way to easily model the vehicle motion is to focus on vehicle kinematics\cite{kine_dyna_4wids}.
Essentially, the difficulty of controlling 4WIDS vehicles comes from the redundancy of the control degrees of freedom, 
which means that there are multiple solutions to achieve a certain vehicle behavior.
The kinematic formulation adds assumptions that the vehicle is moving at a constant velocity, and the tire slip is negligible.
The assumptions can be interpreted as a reduction of the control input space and focusing on considering more practical vehicle motion.

Under the kinematic assumptions, the vehicle's motion can be simplified as a model with three degrees of freedom $\mathbf{u}_{\rm{3DoF}}$ ; 
longitudinal velocity $V_x$, lateral velocity $V_y$, and angular velocity $\omega$ of the vehicle center.
The relationship between the vehicle center velocity $\mathbf{u}_{\rm{3DoF}}$ and the eight-wheel velocities $\mathbf{u}_{\rm{full}}$
 is formulated as follows and shown in Fig.\ref{fig:space_relationship}.

\begin{align}
  &\mathbf{u}_{\rm{3DoF}} = [V_x, V_y, \omega]^T, \\
  &\mathbf{u}_{\rm{full}} = [V_{xfl}, V_{xfr}, V_{xrl}, V_{xrr}, V_{yfl}, V_{yfr}, V_{yrl}, V_{yrr}]^T, \\
  &\mathbf{u}_{\rm{full}} = \mathbf{C}_{\rm{3 \rightarrow 8}} \  \mathbf{u}_{\rm{3DoF}}, \label{eq:vxvyw_to_8dof}\\
  &\mathbf{C}_{\rm{3 \rightarrow 8}} =
  \begin{bmatrix}
    1 & 0 & -d_l \\
    1 & 0 & \ \ d_r \\
    1 & 0 & -d_l \\
    1 & 0 & \ \ d_r \\
    0 & 1 & \ \ l_f \\
    0 & 1 & \ \ l_f \\
    0 & 1 & -l_r \\
    0 & 1 & -l_r \\
  \end{bmatrix}.\label{eq:CA}
\end{align}
The wheel velocity $\mathbf{u}_{\rm{full}}$ can be converted to the vehicle command space $\mathbf{u}_{\rm{vehicle}}$
consisting of four-wheel steering angles $\delta_{*}$ and four-wheel velocities $V_{*}$ for the front left ({\it{fl}}), front right ({\it{fr}}), rear left ({\it{rl}}), and rear right ({\it{rr}}) wheels.
A nonlinear transformation function $f_v$ is used for the conversion,
\begin{align}
  &\mathbf{u}_{\rm{vehicle}} = [\delta_{fl}, \delta_{fr}, \delta_{rl}, \delta_{rr}, V_{fl}, V_{fr}, V_{rl}, V_{rr}]^T, \\
  &\mathbf{u}_{\rm{vehicle}} = f_v(\mathbf{u}_{\rm{full}}), \label{eq:flrr_to_8dof}\\
  &\delta_{*} = \arctan{\frac{V_{y*}}{V_{x*}}},  \\
  &V_{*} = \sqrt{V_{x*}^2 + V_{y*}^2},  \label{eq:fv}
\end{align}
where $*$ represents the wheel positions {\it{fl}}, {\it{fr}}, {\it{rl}}, {\it{rr}}.

Note that the kinematic formulation only requires the geometric relationship of the tires which is easy to obtain, 
and is easily applicable to real-world applications. 

\captionsetup{justification=centering,singlelinecheck=false}

\subsection{Exploring Solution in Redundant Control Input Space}
Sampling based controller needs to get a variety of solutions from the control input space $\mathbf{u}_{\rm{3DoF}}$.
In this study, we found that exploring a slightly redundant space $\mathbf{u}_{\rm{4DoF}}$ and converting it to $\mathbf{u}_{\rm{3DoF}}$
was more effective in achieving smooth and stable vehicle motion with MPPI.
This idea is inspired by methods such as Koopman Operator\cite{koopman_operator} and Dynamic Mode Decomposition\cite{dmd},
which expand the state space to express the complex dynamics as a linear system in a higher dimensional space.

Since $\mathbf{u}_{\rm{4DoF}}$ is a 4-dimensional redundant space, it does not always satisfy kinematic constraints formulated in Eq.(\ref{eq:vxvyw_to_8dof}).
Therefore, projection matrix $\mathbf{C}_{\rm{4 \rightarrow 8}}$ is used to map $\mathbf{u}_{\rm{4DoF}}'$ to $\mathbf{u}_{\rm{vehicle}}$,
so that the sampled vehicle motion always satisfies the kinematic constraints.
\begin{align}
  &\mathbf{u}_{\rm{4DoF}} = [V_{fl}, V_{rr}, \delta_{fl}, \delta_{rr}]^T, \\
  &\mathbf{u}_{\rm{4DoF}}'= f (\mathbf{u}_{\rm{4DoF}}) \nonumber \\
  &\ \ \  = [V_{fl}\cos{\delta_{fl}}, V_{rr}\cos{\delta_{rr}}, V_{fl}\sin{\delta_{fl}}, V_{rr}\sin{\delta_{rr}}]^T \nonumber \\
  &\ \ \  = [V_{xfl}, V_{xrr}, V_{yfl}, V_{yrr}]^T, \\
  &\mathbf{u}_{\rm{vehicle}} = \mathbf{C}_{\rm{4 \rightarrow 8}} \mathbf{u}_{\rm{4DoF}}', \label{eq:flrr_to_8dof}\\
  &\mathbf{C}_{\rm{4 \rightarrow 8}} =
  \mathbf{C_{\rm{3 \rightarrow 8}}} \mathbf{C_{\rm{4 \rightarrow 3}}},\\
  &\mathbf{C_{\rm{4 \rightarrow 3}}} = 
  \begin{bmatrix}
    \frac{d_r}{d_l + d_r} & \frac{d_l}{d_l + d_r} & 0 & 0 \\
    0 & 0 & \frac{l_r}{l_f + l_r} & \frac{l_f}{l_f + l_r}\\
    \frac{-1}{2 (d_l + d_r)} & \frac{1}{2 (d_l + d_r)} & 0 & 0 \\
  \end{bmatrix}. \label{eq:CB}
\end{align}
The conversion to reduce a dimention in Eq.(\ref{eq:CB}) is based on the idea of taking the average of the control inputs between front wheel space ($V_{fl}, \delta_{fl}$) and the rear wheel space ($V_{rr}, \delta_{rr}$)
to obtain the compromised vehicle center angular velocity $\omega$.

\begin{figure*}[t!]
  \centering
  \includegraphics[width=0.85\linewidth]{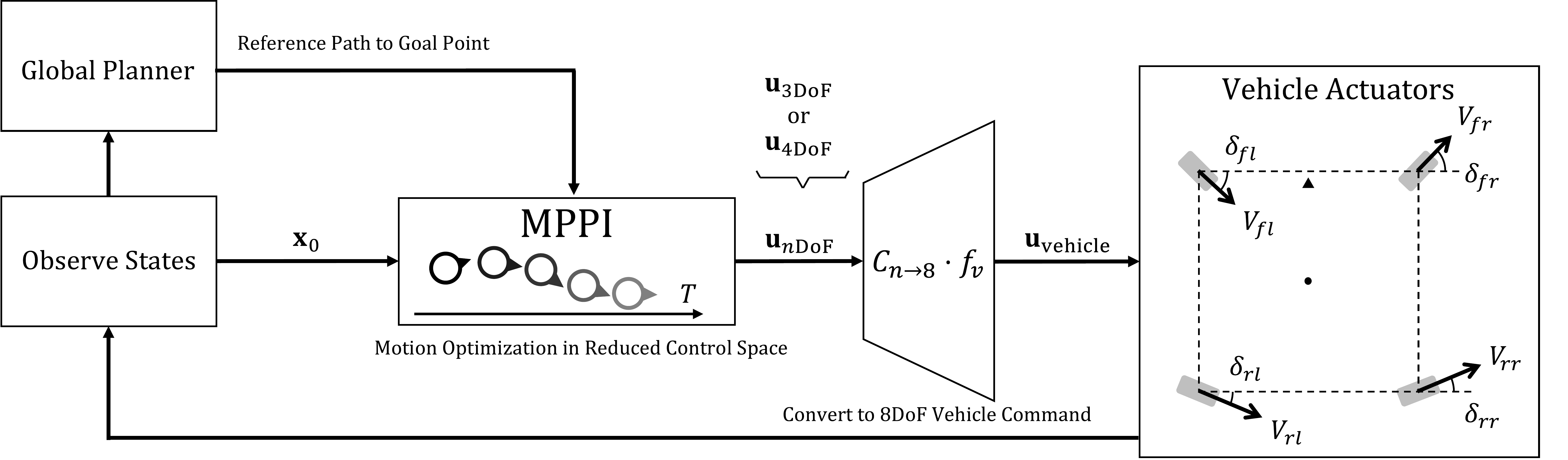}
 \vspace{+0mm}
    \caption{Overview of the control architecture (See Section \ref{sec:system_overview}).
    (a) global planner generates a reference trajectory based on current vehicle pose and the given map,
    (b) MPPI generates the optimal control input in a reduced dimensional space, 
    (c) the $n$ DoF ($n \in \{3, 4 \}$) control input is converted to the 8DoF vehicle command space, 
    and (d) the vehicle actuators execute the command.
    }
 \vspace{+0mm}
  \label{fig:control_architecture}
\end{figure*}

\section{NAVIGATION ARCHITECTURE \\ FOR 4WIDS VEHICLE}

\subsection{System Overview} \label{sec:system_overview}

The navigation system for the 4WIDS vehicle is composed of four main components below in this work (See Fig.\ref{fig:control_architecture})
\subsubsection{State Observation}
The map of the environment is given, and the vehicle localizes itself using 2D LiDAR point cloud data and odometry information.
\subsubsection{Global Path Planning}
The global planner calculates a path from the current vehicle position to the goal using Dijkstra's algorithm.
\subsubsection{Local Path Planning}
In this work, a Model Predictive Path-Integral Controller (MPPI) is used to plan the local path and calculate the next optimal control input.
Since exploring the full vehicle command space which has eight degrees of freedom is difficult and inefficient,
MPPI samples solutions in a dimension that is reasonably reduced by kinematic constraints
such as $\mathbf{u}_{\rm{3DoF}}$ and $\mathbf{u}_{\rm{4DoF}}$ defined in Section \ref{sec:dynamics}.
\subsubsection{Vehicle Command Calculation}
Since MPPI calculates the control input in a reduced dimension, the control input is converted to the 8DoF vehicle command and sent to the vehicle actuators.
The optimal control input calculated by MPPI is converted to the 8DoF vehicle command and sent to the actuators.
The conversion is done by Eq.(\ref{eq:CA}), (\ref{eq:CB}) and (\ref{eq:fv}) in Section \ref{sec:dynamics}. 

\subsection{Algorithm of MPPI Controller}
In this study, local planning is performed using the Model Predictive Path-Integral (MPPI) Controller shown in Algorithm \ref{alg:mppi}.
MPPI is an optimal control algorithm that uses a sample-based approach to compute the optimal control input sequence in the near future.
The algorithm is based on the idea of sampling the control input sequences as normal distributions centered at the previous optimal input sequence, 
and getting the optimal sequence as weighted sum so that better sequences are heavily weighted and vice versa.

For a discrete-time, continuous state-action system $\mathbf{x}_{t} = \mathbf{F}(\mathbf{x}, \mathbf{u})$, 
$K$ samples of input sequences $V = \{ \mathbf{v}_{t} \}_{t=0}^{T-1} $ are generated by adding Gaussian noise 
to mean control input sequence $U = \{ \mathbf{u}_{t} \}_{t=0}^{T-1} $ with covariance matrix $\Sigma$.\  
After preparing the samples, the optimal control input sequence is easily obtained by calculating the weighted sum of the samples.
The stage cost function $c(\mathbf{x}, \mathbf{u})$ and terminal cost function $\phi(\mathbf{x})$ are defined to evaluate the quality of the samples.
Let $S_k$ be the total cost of the $k$-th sample whose input sequence is $V_k$, the weight for the sequence is 
\begin{align}
    & w(V_k) = \frac{1}{\eta} \left( -\frac{1}{\lambda}S(V_k) + \lambda \sum_{\tau=0}^{T-1} \mathbf{u}_{t}^T \Sigma^{-1} \mathbf{v}_{t} - \rho \right)
\end{align}
where $\eta$ is the normalization factor, $\lambda$ is the constant temperature parameter, and $\rho$ is the minimum cost among the samples.

The notable benefits of MPPI are that it can be applied to a wide range of optimization problems, including non-linear, non-convex, and non-differentiable cost functions and system models.
The algorithm is theoretically guaranteed to minimize the forward Kullback-Leibler (KL) divergence 
between the proposed distribution and the optimal distribution which minimizes the total cost function.

\begin{algorithm}[h]
  \caption{Model Predictive Path-Integral Controller}\label{alg:mppi}
  \begin{algorithmic}
  \State \textbf{Given: F}, g: Transition Model;
  \State K: Number of samples;
  \State T: Number of timesteps;
  \State $U \leftarrow (\mathbf{u}_0, \mathbf{u}_1, \dots, \mathbf{u}_{T-1})$: Initial control sequence;
  \State $\Sigma, \phi, c, \gamma, \lambda, \alpha, \Delta t$: Cost functions and parameters;
  \While{task not completed}
  \State $\mathbf{x}_0 \leftarrow \rm{ObserveSystemState}()$
  \For {k = 0 to K-1}
      \State $\mathbf{x} \leftarrow \mathbf{x}_0;$
      \State $\rm{Sample} \ \mathcal{E} = (\epsilon_0^k \dots \mathbf{\epsilon}_{T-1}^k), \ \mathbf{\epsilon} \in \mathcal{N}(0, \Sigma); $
      \For {t = 1 to T-1}
          \If {$k < (1-\alpha) K$}
              \State $\mathbf{v}_{t-1} = \mathbf{u}_{t-1} + \mathbf{\epsilon}^k_{t-1}; $ \Comment{\small samples for exploitation}
          \Else
              \State $\mathbf{v}_{t-1} = \mathbf{\epsilon}^k_{t-1}; $ \Comment{\small samples for exploration}
          \EndIf
          \State $x \leftarrow \mathbf{F}\left(\mathbf{x}, g(\mathbf{v}_{t-1}), \Delta t\right);$
          \State $S_k += c(\mathbf{x}, \mathbf{u}) + \gamma \mathbf{u}_{t-1}^T \Sigma^{-1} \mathbf{v}_{t}$ \Comment{\small add stage cost}
      \EndFor
      \State $S_k += \phi(\mathbf{x})$ \Comment{\small add terminal cost}
  \EndFor
  \State $\rho \leftarrow \min_k [S_k];$ 
  \State $\eta \leftarrow \sum_{k=1}^{K} \exp\left(-\frac{1}{\lambda} (S_k - \rho) \right);$
  \For {k = 0 to K-1}
        \State $\mathbf{w}_k \leftarrow \frac{1}{\eta} \exp\left(-\frac{1}{\lambda} S_k\right);$ \Comment{\small calculate sample weights}
  \EndFor
  \For {t = 0 to T-1}
    \State $U \leftarrow U + \left( \sum_{k=1}^K \mathbf{w}_k \mathcal{E}^k \right);$ \Comment{\small calculate weighted sum}
  \EndFor
  \State $\mathbf{u}_{\rm{vehicle}} \leftarrow \rm{ConvertTo8DoFVehicleCommand}(\mathbf{u}_0);$
  \State $\rm{SendToVehicleActuators}(\mathbf{u}_{\rm{vehicle}});$
  \For {t = 1 to T-1}
      \State $\mathbf{u}_{t-1} \leftarrow \mathbf{u}_t;$ \Comment{\small shift the control sequence}
  \EndFor
  \State $\mathbf{u}_{T-1} \leftarrow \rm{Initialize}(\mathbf{u}_{T-1});$ 
  \EndWhile
  \end{algorithmic}
\end{algorithm}

\subsection{MPPI Switching Multiple Control Spaces}
Through analysis of the experimental results in Section \ref{sec:evaluation_metrics},
we found that selecting the control input space to explore the solution affects the navigation performance significantly.
To take advantage of the strengths and mitigate the weaknesses of each control input space, we propose a method to switch between multiple spaces in real time according to the situation 
explained in Algorithm \ref{alg:hybrid_mppi}.
If two control input spaces are defined as $\mathbf{u}^A$ and $\mathbf{u}^B$, 
both spaces need to update the control input sequence in each time step as a preparation for the next calculation.
The conversion functions $\rm{ConvertToSpaceA}$ and $\rm{ConvertToSpaceB}$ are needed to convert the control input sequence to the other space.
If we switch $\mathbf{u}_{\rm{3DoF}}$ to $\mathbf{u}_{\rm{4DoF}}$, the conversions are done using Eq.(\ref{eq:CA}), (\ref{eq:CB}), and (\ref{eq:fv}) in Section \ref{sec:dynamics}.

\begin{algorithm}[H]
    \caption{MPPI Switching Multiple Control Input Spaces}\label{alg:hybrid_mppi}
        \begin{algorithmic}
        \State $U^A_0 \leftarrow (u_0^A, \dots, u_{T-1}^A)$: Initial control sequence of space A;
        \State $U^B_0 \leftarrow (u_0^B, \dots, u_{T-1}^B)$: Initial control sequence of space B;
        \While{task not completed}
    
        \State mode $\leftarrow \rm{SelectMode}();$ \Comment{\small select control input space}
    
        \If {mode is A}
            \State $U^A_{t+1} \leftarrow \rm{SolveMPPI}(U^A_t);$
            \State $U^B_{t+1} \leftarrow \rm{ConvertToSpaceB}(U^A_{t+1});$
        \ElsIf {mode is B}
            \State $U^B_{t+1} \leftarrow \rm{SolveMPPI}(U^B_t);$
            \State $U^A_{t+1} \leftarrow \rm{ConvertToSpaceA}(U^B_{t+1});$
        \EndIf
    
        \EndWhile
        \end{algorithmic}
      \end{algorithm}

\section{EXPERIMENTS}

\begin{table*}[t]
  \caption{Evaluation Results of 100 Navigation Episodes \\ \textcolor{blue}{blue value is the best score}, and \textcolor{red}{red value is the worst score} of all four controllers.}
  \vspace{-1mm}
  \label{tab:simulation_result}
  \centering
  \begin{tabular}{ c | cccc | cccc }
  \toprule
  Field & \multicolumn{4}{c|}{Cylinder Garden} & \multicolumn{4}{c}{Maze} \\
  \midrule
  Method & \multirow{2}{*}{\begin{tabular}[c]{@{}c@{}}MPPI-3D(a)\\ $[V_x, V_y, \omega]$\end{tabular}}
         & \multirow{2}{*}{\begin{tabular}[c]{@{}c@{}}MPPI-3D(b)\\ $[V_x, V_y, \omega]$\end{tabular}}
         & \multirow{2}{*}{\begin{tabular}[c]{@{}c@{}}MPPI-4D\\ $[V_{fl}, V_{rr}, \delta_{fl}, \delta_{rr}]$\end{tabular}} 
         & \multirow{2}{*}{\begin{tabular}[c]{@{}c@{}}MPPI-H\\ {\scriptsize 3D(a) / 4D} \end{tabular}}

         & \multirow{2}{*}{\begin{tabular}[c]{@{}c@{}}MPPI-3D(a)\\ $[V_x, V_y, \omega]$\end{tabular}}
         & \multirow{2}{*}{\begin{tabular}[c]{@{}c@{}}MPPI-3D(b)\\ $[V_x, V_y, \omega]$\end{tabular}}
         & \multirow{2}{*}{\begin{tabular}[c]{@{}c@{}}MPPI-4D\\ $[V_{fl}, V_{rr}, \delta_{fl}, \delta_{rr}]$\end{tabular}} 
         & \multirow{2}{*}{\begin{tabular}[c]{@{}c@{}}MPPI-H\\ {\scriptsize 3D(b) / 4D} \end{tabular}}
         \\
  Control Space &  &  &  &  &  &  &\\
  \midrule
    Cost {[}-{]} $\downarrow$ & \textcolor{red}{3241.7} & 1900.5 & \textcolor{blue}{1455.8} & 2425.4 & \textcolor{red}{10030.4} & 3918.8 & \textcolor{blue}{2452.3} & 2887.6 \\
    Calc. Time {[}ms{]} $\downarrow$ & 24.1 & \textcolor{blue}{23.0} & \textcolor{red}{27.6} & 26.6 & \textcolor{blue}{19.7} & 19.9 & \textcolor{red}{24.0} & 21.0 \\
    Steering Rate {[}rad/s{]} $\downarrow$ & \textcolor{red}{4.5} & \textcolor{blue}{3.1} & 3.6 & 4.0 & \textcolor{red}{6.0} & 3.6 & 5.0 & \textcolor{blue}{3.5} \\
    Wheel Acc. {[}$\rm{m/s^2}${]} $\downarrow$ & \textcolor{red}{5.03} & \textcolor{blue}{3.36} & 4.08 & 4.98 &  \textcolor{red}{6.02} &  \textcolor{blue}{3.77} & 4.85 & 4.02 \\
    Trajectory Length {[}m{]} $\downarrow$ & \textcolor{red}{51.9} & 46.0 & \textcolor{blue}{40.8} & 42.6 & \textcolor{red}{72.1} & 64.8 & \textcolor{blue}{55.2} & 55.3 \\
    Episode Time {[}s{]} $\downarrow$ & 36.4 & \textcolor{red}{41.3} & 38.4 & \textcolor{blue}{31.2} & 49.6 & \textcolor{red}{55.9} & 52.1 & \textcolor{blue}{44.8} \\
    Success Rate {[}\%{]} $\uparrow$ & \textcolor{red}{76} & 89 & \textcolor{blue}{100} & 99 & \textcolor{red}{33} & 58 & \textcolor{blue}{98} & 96 \\
  \bottomrule
  \end{tabular}
  \vspace{-3mm}
\end{table*}

\subsection{Simulation Setup}
\begin{figure}[t]
  \begin{minipage}[b]{0.47\linewidth}
    \centering
    \includegraphics[keepaspectratio, width=0.9\linewidth]{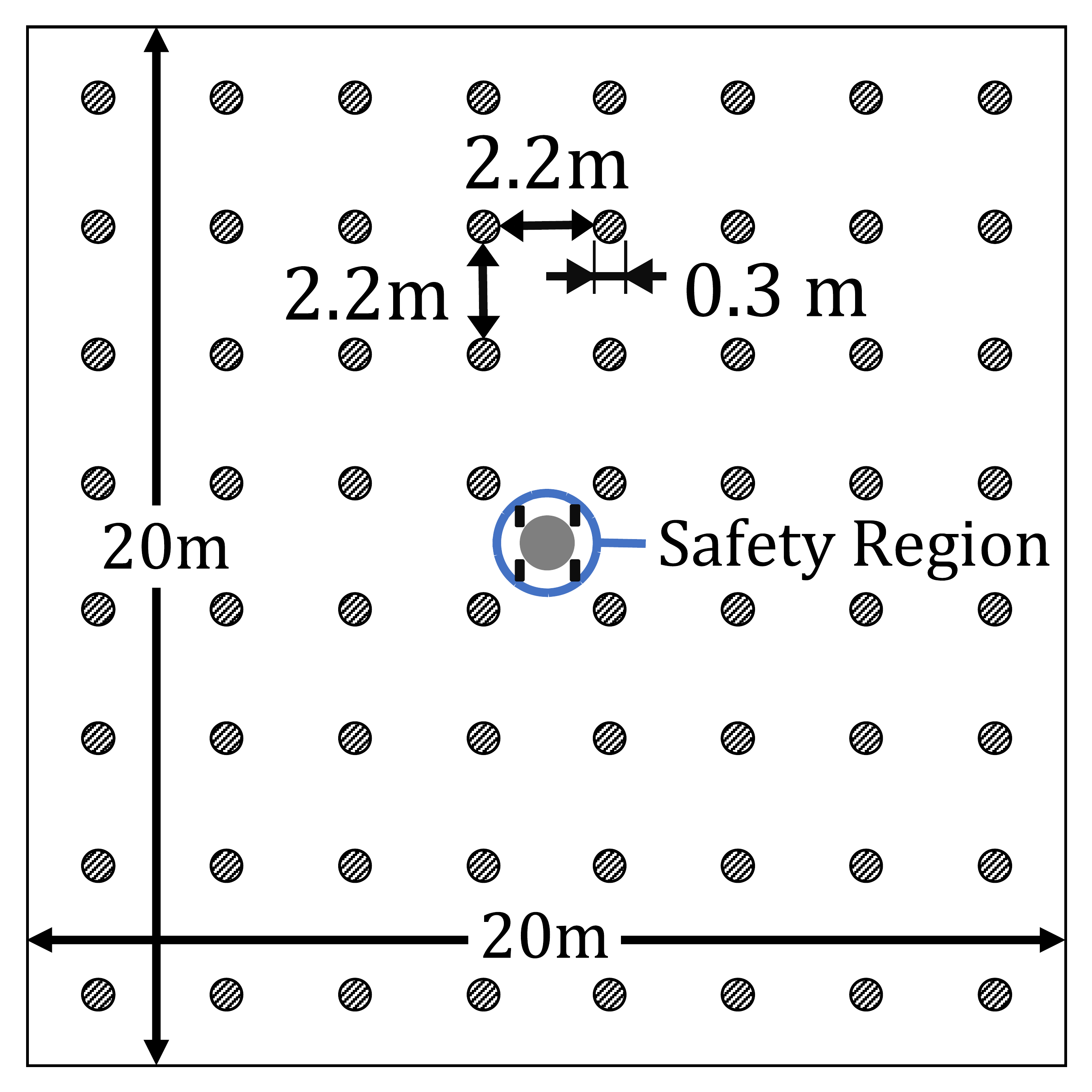}
    \subcaption{Cylinder Garden (Easier)}
  \end{minipage}
  \begin{minipage}[b]{0.47\linewidth}
    \centering
    \includegraphics[keepaspectratio, width=0.9\linewidth]{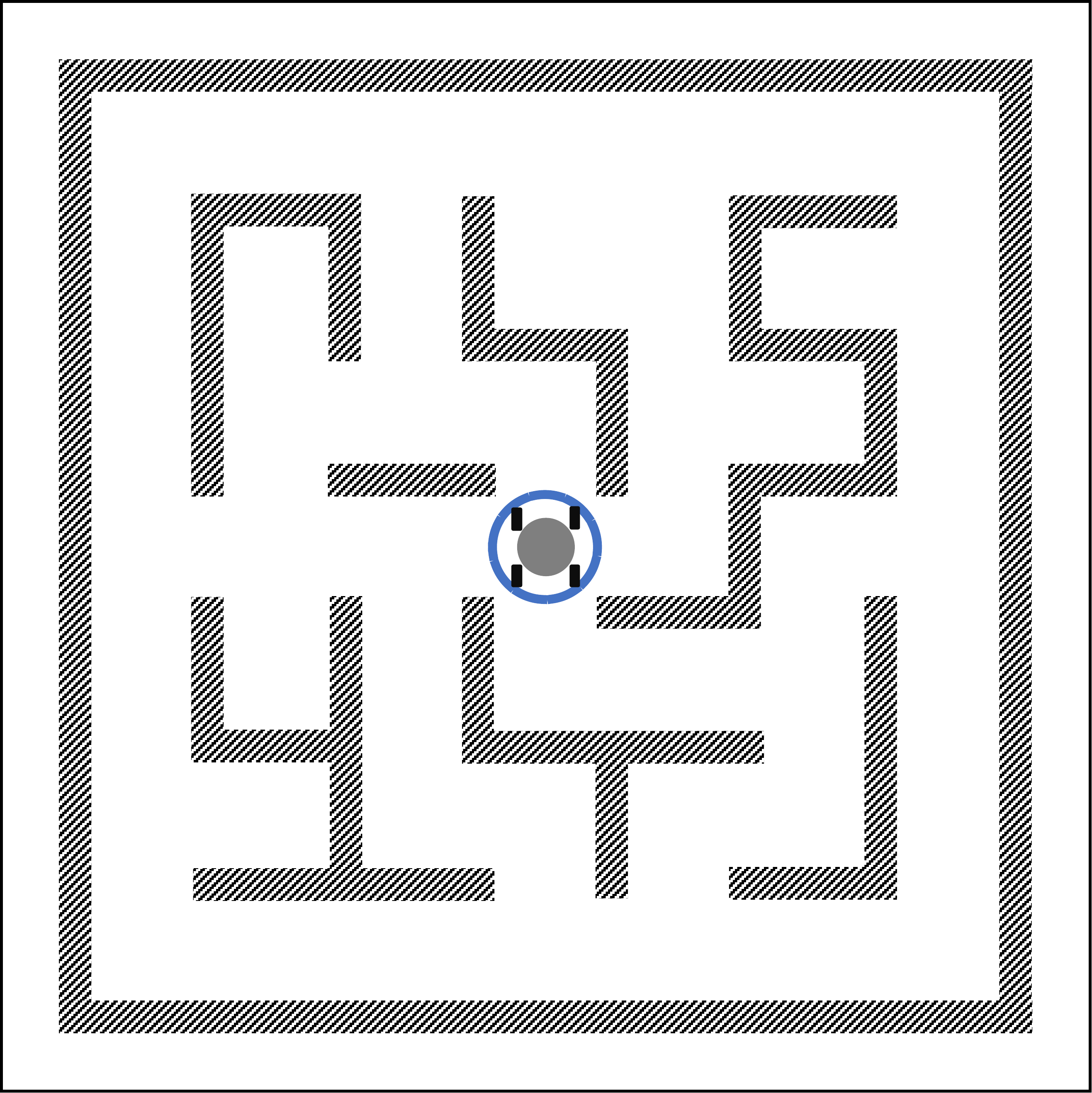}
    \subcaption{Maze (Harder)}
  \end{minipage}
  \caption{Simulation environment}
  \label{fig:sim_env}
\end{figure}

To evaluate the navigation performance of the proposed architecture, 
we set up a simulation environment using Gazebo simulator. 
Two types of fields are prepared, "Cylinder Garden" and "Maze" (Fig. \ref{fig:sim_env}).
The vehicle need to reach 10 goals sequentially in a episode as fast as possible while avoiding densely placed obstacles.
The four wheel positions are set symmetrically as $l_f = l_r = d_l = d_r = 0.5 \ \rm{[m]}$ (Fig. \ref{fig:space_relationship}).

Our evaluation system was developed using ROS and C++.
Calculation is performed on a desktop computer with an Intel Core i7-13700KF CPU and 32GB of RAM.
For faster computation, we used CPU multi-threading with OpenMP\cite{openmp}.

\subsection{Cost Formulation for MPPI}

In this section, the cost functions $c(\mathbf{x}, \mathbf{u})$ and $\phi(\mathbf{x})$ (See Algorithm \ref{alg:mppi} are defined for the MPPI controller.

The stage cost $c(\mathbf{x}, \mathbf{u})$ is defined as the weighted sum of the following terms,
\begin{align}
    c(\mathbf{x}, \mathbf{u}) &= 40\ c_{\rm{dist}}(p_x, p_y) + 30\ c_{\rm{angle}}(\theta) +  10\ c_{\rm{speed}}(v) \nonumber \\ 
    &\ + 50\ c_{\rm{collision}}(p_x, p_y) + c_{\rm{cmd}}(\mathbf{u}),
\end{align}
where $p_x$ and $p_y$ are the vehicle's position, $\theta$ is the vehicle's yaw angle, and $v$ is the vehicle's velocity.
$c_{\rm{dist}}(p_x, p_y)$, $c_{\rm{angle}}(\theta)$, and $c_{\rm{speed}}(v)$ are quadratic error from the reference path and constant target velocity $v_{\rm{des}}$, respectively.
$c_{\rm{collision}}(p_x, p_y)$ is a binary cost function that returns 1 if the vehicle is in collision, and 0 otherwise;
\begin{align}
  c_{\rm{collision}}(p_x, p_y) = \begin{cases}
    0 & \text{if no collision} \\
    1 & \text{if collision}
  \end{cases}.
\end{align}
$c_{\rm{cmd}}(\mathbf{u})$ is used to smooth the vehicle actuator commands.
In either control space $\mathbf{u}_{\rm{3DoF}}$ or $\mathbf{u}_{\rm{4DoF}}$ to be explored, it can be converted to 8-dimensional vehicle actuator commands $\mathbf{u}_{\rm{vehicle}}$ with 
equations Eq.\eqref{eq:vxvyw_to_8dof} and \eqref{eq:flrr_to_8dof}.
With previous vehicle command $\mathbf{u}_{\rm{vehicle}}^{\rm{prev}}$ in the control sequence, minimizing the penalty term
\begin{align}
  c_{\rm{cmd}}(\mathbf{u}) = \| \mathbf{u}_{\rm{vehicle}} - \mathbf{u}_{\rm{vehicle}}^{\rm{prev}} \|_2
\end{align}
can smooth the vehicle actuator commands.

the terminal cost $\phi(\mathbf{x})$ is added only for preventing reverse driving,
\begin{align}
  \phi(\mathbf{x}) = 50 \  \phi_{\rm{goal}}(p_x, p_y),
\end{align}
where $\phi_{\rm{goal}}(p_x, p_y)$ is the quadratic error from the goal position.

\subsection{Preparing MPPI Controllers for Comparison}

To investigate the characteristics of the MPPI controller depending on the choice of the control space,
MPPI-3D and MPPI-4D are prepared to explore two different control spaces, $\mathbf{u}_{\rm{3DoF}}$ and $\mathbf{u}_{\rm{4DoF}}$.

Since the variance parameter affects the controller's behavior, setting the variance of the control space carefully is important for fair comparison.
Changing the variance parameters, two types of MPPI-3D are prepared, MPPI-3D(a) and MPPI-3D(b).
The variance parameters of MPPI-3D(a) and MPPI-4D are set systematically as shown in Table \ref{tab:mppi_variance} following the rule that 
the variance is half of the maximum value of the control space defined in Table \ref{tab:controller_params}.
This consideration is to explore a wide range of the control space, as the normal distribution contains about 95\% of the data within twice the standard deviation.

Another approach is to fit a normal distribution numerically close to the sampled results of MPPI-4D.
MPPI-3D(b) follows this procedure, and the variance parameters are calculated with maximum likelihood estimation of the normal distribution,
when the vehicle is stopped and all the steering angles are zero in the space of $\mathbf{u}_{\rm{4DoF}}$.

Additionally, a hybrid MPPI controller, MPPI-H, is prepared to switch between MPPI-3D and MPPI-4D depending on the situation.
From the verification results, we determined that the mode selection should be based on the target path tracking error.
The mode switching function in Algorithm \ref{alg:hybrid_mppi} is defined as follows with constant parameter $d_{\rm{thresh}} = 0.3 \ \rm{[m]}$ and $\theta_{\rm{thresh}} = 0.3 \ \rm{[rad]}$,
\begin{align}
  &\rm{SelectMode}() \nonumber \\
  &=\begin{cases}
    \considerhere{ココ} \mathbf{u}_{\rm{3DoF}} & \text{if } c_{\rm{dist}}(p_x, p_y) < d_{\rm{thresh}} 
    \ \rm{and} \  c_{\rm{angle}}(\theta) < \theta_{\rm{thresh}}   
    \\
    \considerhere{ココ} \mathbf{u}_{\rm{4DoF}} & \text{otherwise}
  \end{cases}.
\end{align}

\begin{table}[t]
  \centering
  \begin{minipage}[t]{.4\linewidth}
    \caption{MPPI Params}
    \begin{tabular}{ c | c c }
      \toprule
      Param & Value & Unit \\
      \midrule
      K & 3000 & sample \\
      T & 30 & step \\
      $\Delta t$  & 0.033 & sec \\
      $\alpha$ &  0.1 & - \\
      $\lambda$ & 250 & - \\
      $\gamma$ & 6.25 & - \\
      \bottomrule
      \end{tabular}
    \label{tab:mppi_common_params}
  \end{minipage}
  \hfill
  \begin{minipage}[t]{.55\linewidth}
    \caption{Controller Params}
    \begin{tabular}{ c | c c }
      \toprule
      Param & Value & Unit \\
      \midrule
      Control Interval $\Delta t_i$ & 0.05 & sec \\
      Target Velocity $v_{\rm{des}}$ & 2.00 & m/s \\
      Max. Velocity $v_{\rm{max}}$ & 2.00 & m/s \\
      Max. Yawrate $\omega_{\rm{max}}$ & 1.58 & rad/s \\
      Max. Steering Angle & 1.58 & rad \\
      \bottomrule
      \end{tabular}
    \label{tab:controller_params}
  \end{minipage}
\end{table}

\begin{table}[t]
  \caption{MPPI Variance Params}
  \vspace{-1mm}
  \label{tab:mppi_variance}
  \centering
  \begin{tabular}{ c | c c }
  \toprule
  Name & Control Space & Variance $\Sigma$ \\
  \midrule
  MPPI-3D(a) & $[V_x, V_y, \omega]$ & $[1.00, 1.00, 0.78]$ \\
  MPPI-3D(b) & $[V_x, V_y, \omega]$ & $[0.55, 0.55, 0.96]$ \\
  MPPI-4D & $[V_{fl}, V_{rr}, \delta_{fl}, \delta_{rr}]$ & $[1.00, 1.00, 0.78, 0.78]$ \\
  \bottomrule
  \end{tabular}
\end{table}

\subsection{Definition of Evaluatioin Metrics} \label{sec:evaluation_metrics}

Here the evaluation metrics are defined to compare the performance of the MPPI controllers.
All metrics are calculated for all episodes, and the mean value is used for evaluation.

\subsubsection{Cost}
"Cost" is the sum of stage cost and terminal cost of optimal trajectory output from MPPI.
This metric indicates how well the MPPI controller can minimize the cost function and get close to the optimal behavior.

\subsubsection{Vehicle Command Change}

To evaluate the smoothness of the vehicle actuator commands, two metrics are defined.
"Steering Rate" is the absolute value of the steering angular velocity.
"Wheel Acceleration" means the absolute change of the wheel velocity.
For both metrics, the mean values of four wheels are used for evaluation.

\subsubsection{Navigation Efficiency}
two metrics are defined to evaluate the navigation efficiency.
"Trajectory Length" is selected to evaluate how the vehicle could reach the goal with a short path.
"Episode Time" is also used to know how fast the vehicle could reach the goal.

\subsubsection{Success Rate}
Success rate is the percentage of episodes that the vehicle reached all the given goal points.
Collision with obstacles and getting stuck in the field are major factors of failure.

\subsection{Evaluation Results Comparison} \label{sec:evaluation_results}

100 episodes of navigation are performed for each field, and the evaluation results are shown in Table \ref{tab:simulation_result}.
For each MPPI controller, the mean calculation time is less than 30ms and works in real-time with the control interval $\Delta t_i = 50 \ \rm{[ms]}$.
Even though the cost function and algorithmic parameters are common, the results drastically changed by exploring different control spaces.

MPPI-3D(a) has a short episode time, which means it can drive efficiently.
A lower success rate (76\% in Cylinder Garden, 33\% in Maze) means that it fails to complete episodes frequently, and has difficulty in safe and stable navigation.
Since MPPI-3D(b) has a smaller variance in the vehicle velocity space than MPPI-3D(a), it moves slowly and appears more conservative behavior.
The episode time is the worst among the four controllers in both fields.
However, the success rate is only slightly improved (89\% in Cylinder Garden, 58\% in Maze) from MPPI-3D(a) and still has low stability.

On the other hand, MPPI-4D showed an extremely high success rate (100\% in Cylinder Garden, 99\% in Maze) and stable navigation behavior.
Cost and trajectory length are the best among the four controllers, it means that MPPI-4D is good at finding the optimal solution.
Comparing the episode time in Cylinder Garden, MPPI-4D (38.4s) is faster than MPPI-3D(b) (41.3s), but MPPI-3D(a) (36.4s) surpasses MPPI-4D in terms of driving efficiency.

Summarizing the characteristics of MPPI-3D and MPPI-4D, MPPI-3D is good at high-speed driving but has low stability, and MPPI-4D is more conservative and is good at stable navigation but has lower efficiency.
Then it is reasonable to switch control spaces depending on the situation to balance the efficiency and stability.
In the case the vehicle is in a difficult situation (i.e. the target path tracking error is large), MPPI-4D is selected to ensure stable navigation.
Otherwise, MPPI-3D is selected to drive efficiently.

As a result, the hybrid MPPI-H switching control spaces showed the best episode time (31.2s in Cylinder Garden, 44.8s in Maze), keeping the high success rate (99\% in Cylinder Garden, 96\% in Maze) 
compared to MPPI-4D.
The fact that MPPI-H has no worst score in any of the metrics is also evidence that MPPI-H has a balanced performance.

\subsection{Trajectory Comparison}

\begin{figure}[t]
  \centering
  \includegraphics[keepaspectratio, width=0.99\linewidth]{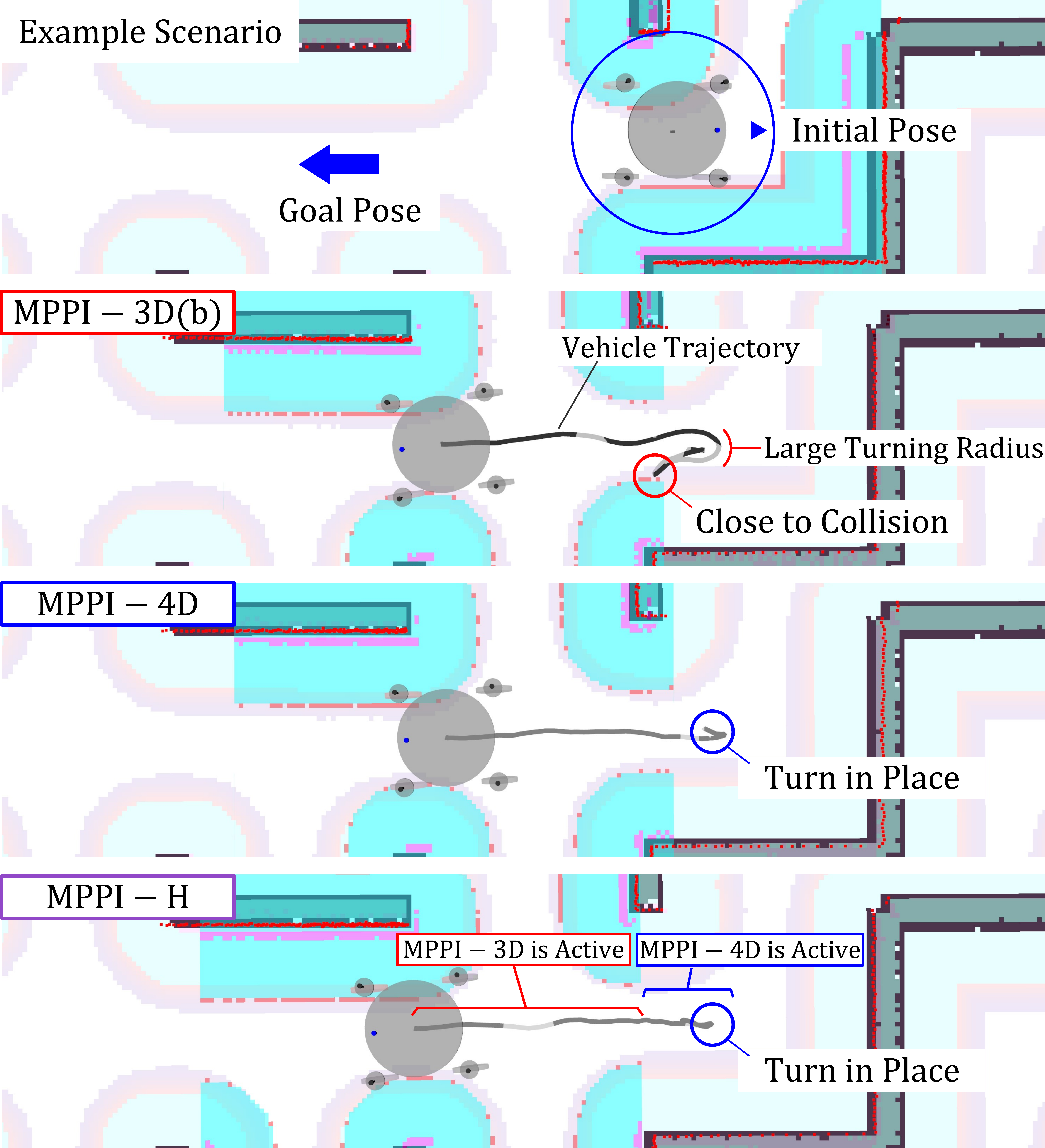}
  \caption{Trajectory comparison of a difficult situation. \\
  While MPPI-3D shows dangerous behavior close to the collision, MPPI-4D and MPPI-H drive safely with turning in a small radius.
  }
  \label{fig:traj_comparison}
\end{figure}

To understand the characteristics of the controllers more deeply, the trajectories of the four controllers are compared in Fig. \ref{fig:traj_comparison}.
This example scenario is a hard situation where the vehicle receives the next goal point and needs to turn sharply.

In Fig. \ref{fig:traj_comparison}, the MPPI-3D(b) turns with a larger radius than the other controllers, 
and a part of the trajectory is close to the collision with surrounding obstacles.
It is a typical dangerous behavior and shows the reason of the lower success rate of MPPI-3D controllers.

On the other hand, MPPI-4D can turn in a small radius and run safely keeping a distance from the obstacles,
showing the reason of the high success rate of MPPI-4D.

In the situation where the vehicle should turn sharply, MPPI-H activates MPPI-4D to drive carefully, resulting in the same turning behavior as MPPI-4D.

\captionsetup[figure]{justification=raggedright}

\section{DISCUSSION}

\captionsetup[figure]{justification=raggedright}
\begin{figure}[t]
  \centering
  \includegraphics[keepaspectratio, width=0.99\linewidth]{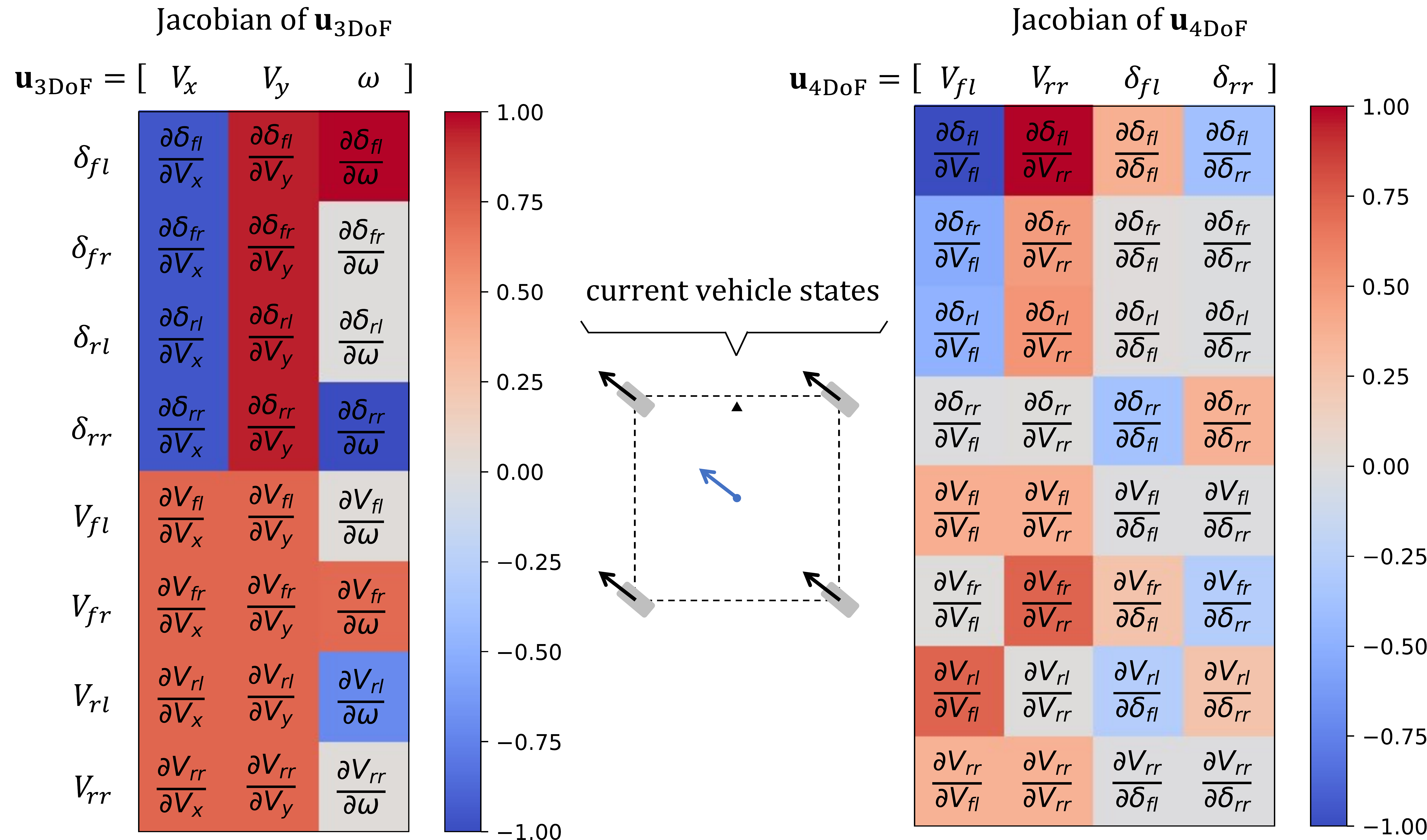}
  \caption{Jacobian matrices comparison between $\mathbf{u}_{\rm{3DoF}}$ and $\mathbf{u}_{\rm{4DoF}}$ \\
  when $\delta_{fl} = \delta_{fr} = \delta_{rl} = \delta_{rr} = 0.25\pi \ [\rm{rad}]$, \\
  $\ \ \ \ \ \ \ \  V_{fl} = V_{fr} = V_{rl} = V_{rr} = 0.7 \ [\rm{m/s}]$.\\
  $J_B$ has a more sparse structure than $J_A$, which makes the optimal solution search easier.
  }
  \label{fig:jacobian}
\end{figure}

In this section,  we discuss why the performance changes by selecting control input space for sampling.
Specifically, explanation of why the solution search in the control space $\mathbf{u}_{\rm{4DoF}}$ is more likely to find a more optimal solution than in $\mathbf{u}_{\rm{3DoF}}$ is provided.

In the navigation task, one of the most difficult situation is when the next goal point is specified in the opposite direction to the vehicle's heading.
In this case, the optimal behavior is to quickly decelerate the vehicle and rotate the vehicle body to the target direction.

Here we consider the Jacobian matrices $J_A, J_B$ with respect for the projection from each control input space $\mathbf{u}_{\rm{3DoF}}, \mathbf{u}_{\rm{4DoF}}$ to the vehicle command space $\mathbf{u}_{\rm{vehicle}}$.
The Jacobian matrices are normalized to the range of -1 to 1 and plotted as color maps in Fig.\ref{fig:jacobian}.

As for the control input space $\mathbf{u}_{\rm{3DoF}}$, deceleration of the vehicle speed $V_x$ is strongly affects the vehicle steering angles.
Therefore, even if a control sequence that includes a rapid deceleration is sampled, it is likely to be penalized due to the large change in the steering angle.
As a result, the deceleration behavior is less likely to occur
and the vehicle tries to rotate with the current speed, which sometimes causes collision with obstacles.

On the other hand, as for the control input space $\mathbf{u}_{\rm{4DoF}}$, the Jacobian matrix $J_B$ has a more sparse structure than $J_A$.
This means that the change in the vehicle speed and the steering angles are relatively independent.
This feature makes the optimal solution search easier, and more likely to find the better solution including 
rapid deceleration and rotation of the vehicle at the same time.

From the above discussion, it can be considered that the sparsity of the Jacobian matrix is one of the guidelines for selecting the control input space for MPPI.

\section{CONCLUSION}
This paper presented a navigation framework for four-wheel independent driving and steering (4WIDS) Vehicle based on the Model Predictive Path-Integral (MPPI) control method.
For efficiently solving the optimal control problem for high-dimensional systems, 
two types of reasonably reduced control spaces are introduced.
Evaluation results show that the control space which has a slightly redundant dimension than the bare minimum can achieve more stable navigation.
In addition, our novel approach that switches the control space in real-time can achieve both efficiency and stability at a high level.

Although this work offered a new perspective on how to choose an effective sampling space, 
devising a systematic method to find it is the next step.
In addition, implementing the proposed method on a real platform is needed to verify its effectiveness in real-world environments.

\vspace{5mm}

\addtolength{\textheight}{-12cm}   



\bibliographystyle{IEEEtran}
\bibliography{reference}

\end{document}